\documentclass[a4paper,12pt]{article}

\usepackage{amsmath,amssymb,amsfonts}
\usepackage{graphicx}
\usepackage{booktabs}
\usepackage{hyperref}
\usepackage{cleveref}
\usepackage{xcolor}
\usepackage{array}
\usepackage{subcaption}
\usepackage{multirow}
\usepackage{float}
\usepackage[margin=1in]{geometry}
\newcommand{\noise}{\mathbf{z}}
\newcommand{\context}{\mathbf{x}}

\title{\textbf{Noise-Space Attribution and Control of Chunk-Boundary Artifact}}
\author{Rui Wang\textsuperscript{1}}
\date{}

\begin{document}
\maketitle
\footnotetext[1]{Nanjing University}

\begin{abstract}
Action chunking is widely used in generative visuomotor policies, yet the recurring execution discontinuities at chunk boundaries still lack a mechanistic explanation. This paper treats chunk-boundary artifact as an analyzable mechanism variable. We first show that successful and failed episodes separate stably on artifact metrics. We then show that, in stochastic action-chunked policies, fixing the observation context and changing only latent noise is sufficient to modulate artifact systematically. On the same Diffusion Policy checkpoint, comparisons among DDPM, zero-variance DDPM, and DDIM further show that this local controllability depends on whether the information path from initial noise to action output remains intact. Finally, from controlled interventions at fixed local execution states, we find that artifact changes can carry through to final outcome, and that the preferred direction can reverse even within the same task: some contexts achieve higher success under lower artifact, whereas others achieve higher success under higher artifact. In a representative high-artifact-favoring key context selected by held-out matched-continuation validation, success rate increases from $0.033$ to $0.717$. These results show that chunk-boundary artifact is not a mere execution-side by-product, but a variable in noise space that can be attributed, controlled, and mechanistically linked to task outcome.
\end{abstract}

\section{Introduction}
\label{sec:intro_en_v3}

Action chunking---predicting a short sequence of future actions in one forward pass and executing only part of it---has become a standard design pattern in generative robot policies~\cite{chi2023diffusion,zhao2023aloha,black2024pi0}. This design improves inference efficiency by reducing replanning frequency and often yields smooth motion \emph{within} a chunk. However, when one chunk ends and the next chunk is replanned, the executed trajectory can exhibit recurring discontinuities at chunk boundaries. In practice, this phenomenon is often treated as an engineering issue to be smoothed or repaired, rather than as a mechanism variable that merits direct explanation.

Most prior work treats boundary discontinuity as an execution-side by-product and focuses on post hoc remedies such as temporal ensembling, cross-chunk continuation modeling, or re-execution control~\cite{zhao2023aloha,learning_native_continuation,rtc2024,remac2025,liu2025bid}. These methods can mitigate the boundary problem, but they leave a more basic question unanswered:
\emph{what determines the strength of chunk-boundary artifact, and can this mechanism be directionally intervened on in noise space so as to influence task outcome?}

Our basic finding is that, for stochastic action-chunked policies, noise space contains structures that can stably drive chunk-boundary artifact, and that in some contexts these artifact changes continue to propagate to task outcome. The focus here is therefore not another boundary-smoothing technique, but the generation of boundary artifact itself, its attribution to latent noise, and its relationship to outcome.

Concretely, the paper does three things:
\begin{enumerate}
  \item \textbf{Mechanism-variable formulation.} We introduce both episode-level and local metrics for chunk-boundary artifact and thereby formulate it as an analyzable mechanism variable.
  \item \textbf{Evidence for noise-space attribution.} We show that under fixed observation contexts, changing only latent noise is sufficient to systematically modulate boundary artifact, and that this controllability depends on whether an intact information path exists from initial noise to action output.
  \item \textbf{Outcome-level bridge.} Through controlled interventions, we show that changes in boundary artifact can propagate to task outcome in some settings, and that this propagation is strongly context-dependent.
\end{enumerate}

\section{Related Work}
\label{sec:related_en_v3}

\paragraph{Action chunking and boundary discontinuity.}
ACT~\cite{zhao2023aloha}, RTC~\cite{rtc2024}, REMAC~\cite{remac2025}, and native continuation~\cite{learning_native_continuation} all acknowledge that chunk boundaries are a special regime. Most of this work, however, is about making the boundary smoother---through temporal ensembling, re-execution, or continuation modeling. Our question is earlier in the chain: why does boundary artifact appear at all, and can it already be analyzed and controlled in noise space before execution?

\paragraph{Structure in generative-model noise spaces.}
InitNO~\cite{initno2024} and NoiseCLR~\cite{dalva2024noiseclr} suggest that noise space itself can carry exploitable semantic structure. We bring that perspective into generative robot control. The question here is not whether noise contains some abstract structure in general, but which directions in noise space show up concretely as chunk-boundary artifact and then matter for outcome.

\paragraph{Latent-level policy improvement.}
Methods such as DSRL~\cite{dsrl2024} study how latent manipulations can increase reward, typically through online interaction and reward signals. This paper is not another latent optimization method. Its purpose is to explain why latent intervention changes success at all. The difference is that prior work mainly treats noise as a means of policy improvement, whereas we treat artifact as the mechanism to be explained. Our analysis also does not rely on an additional online reward-optimization loop.

\section{Preliminaries and Artifact Definition}
\label{sec:prelim_en_v3}

\subsection{Action-chunked generative policy}

We consider a visuomotor policy $\pi_\theta$ that receives an observation context $\context_t$ at time step $t$ and, for stochastic policies, latent noise $\noise$, and generates a future action chunk:
\begin{equation}
  \mathbf{a}_{t:t+H} = \pi_\theta(\context_t, \noise).
\end{equation}
At deployment time, the agent executes only the first $K \le H$ actions and then replans from the new context $\context_{t+K}$. As a result, the trajectory periodically switches chunks at replanning boundaries $t, t+K, t+2K, \ldots$.

\subsection{Boundary-artifact metrics}

We measure action discontinuity at the boundary using action jerk:
\begin{equation}
  j_t = \left\lVert \mathbf{a}_t - 2\mathbf{a}_{t-1} + \mathbf{a}_{t-2} \right\rVert_2.
\end{equation}
Let $p(t)=t \bmod K$ denote the phase within a replanning cycle, with boundary phases $\mathcal{B}=\{0,1\}$ and interior phases $\mathcal{I}=\{2,3,4\}$, and let $\bar{j}_p$ denote the mean of $j_t$ over all timesteps in the episode whose phase equals $p$. We define the episode-level boundary--interior jerk contrast as
\begin{equation}
  J_{\mathrm{contrast}}=
  \frac{1}{|\mathcal{B}|}\sum_{p\in\mathcal{B}}\bar{j}_p-
  \frac{1}{|\mathcal{I}|}\sum_{p\in\mathcal{I}}\bar{j}_p.
\end{equation}

To study local mechanism changes, we further define first-boundary metrics around the first executable replanning boundary. We focus on the first boundary because later ones are progressively contaminated by state drift, contact events, and compounding replanning error, making the effect of noise itself harder to isolate. The first-boundary transition jerk (BTJ) is the first jump at the boundary itself. The first-boundary gap is defined as the mean of the first two local boundary jerks minus the mean of the subsequent interior jerks. Episode-level metrics describe boundary discontinuity over a whole rollout, whereas local metrics characterize the mechanism at a single replanning transition.

\subsection{Experimental setup and mechanism-chain overview}

Unless otherwise stated, we use frozen OpenPI flow-matching checkpoints. The main experiments use \texttt{pi0.5-libero}, while a small number of near-saturation controls use \texttt{pi0-libero}. Goal 0 (opening the middle drawer) and goal 3 (opening the top drawer and placing the bowl inside) are the two LIBERO~\cite{libero2023} goals used to compare artifact separation between successful and failed episodes; task 8 is the representative non-saturated task used for outcome-level steering analysis.

In fixed-context experiments, we refer to the local execution state formed at a specified replanning boundary of a rollout prefix as a \emph{key context}, written as $(seed, trial, replan)$ for rollout seed, episode index under that seed, and boundary index within the episode.

In outcome-level intervention experiments, we first fix the reference noise at the boundary and then search for local steering directions around it. A direction that lowers the local artifact at that boundary is denoted target-low, and one that raises it is denoted target-high. The \emph{reference gap} is the first-boundary gap produced by the reference noise itself, and can be understood as the baseline artifact level of that context. When we report results ``after subtracting the reference gap,'' we mean that this context-specific baseline has been removed before comparing the additional variation induced by different noises.

The experimental chain in this paper addresses three questions. First, under a fixed context, is changing noise alone sufficient to change artifact? Second, does this controllability depend on whether the sampler preserves an intact information path from initial noise to action output? Third, when we intervene along artifact-aligned directions, can these local changes further propagate to final outcome? The next three sections correspond to these three links in the mechanism chain.

\section{Artifact as a Mechanism Variable: Separation First, Control Second}
\label{sec:artifact_variable_en_v3}

We first establish why chunk-boundary artifact is worth treating as a mechanism variable, and then show that it can be controlled stably in noise space.

\subsection{Artifact separation between successful and failed episodes}

We begin with a basic question: is artifact genuinely related to task outcome? \Cref{tab:artifact_outcome_en_v3} reports rollout statistics for the frozen OpenPI main checkpoint (\texttt{pi0.5-libero}). On both goal 3 and goal 0, failed episodes exhibit substantially stronger episode-level boundary--interior jerk contrast than successful episodes. More importantly, this separation remains under matched-horizon and contact-free controls, indicating that it cannot be reduced to trivial explanations such as longer failed episodes, more contact events, or terminal instability. Chunk-boundary artifact is therefore not merely an execution-side nuisance, but a mechanism variable stably associated with failure risk.

\begin{table}[t]
  \centering
  \caption{Episode-level boundary--interior jerk contrast for successful and failed rollouts. All $p$ values are from 20,000-sample permutation tests.}
  \label{tab:artifact_outcome_en_v3}
  \begin{tabular}{llcccc}
    \toprule
    Task & Control condition & Success mean & Failure mean & $\Delta$ & $p$ value \\
    \midrule
    \multirow{3}{*}{goal 3}
      & All episodes & 0.102 & 0.341 & +0.238 & $<10^{-4}$ \\
      & Contact-free & 0.100 & 0.297 & +0.197 & $<10^{-4}$ \\
      & Contact-free first 50 steps & 0.093 & 0.138 & +0.045 & 0.005 \\
    \midrule
    \multirow{3}{*}{goal 0}
      & All episodes & 0.108 & 0.219 & +0.110 & $<10^{-4}$ \\
      & Contact-free & 0.108 & 0.185 & +0.077 & 0.002 \\
      & Contact-free first 50 steps & 0.049 & 0.070 & +0.021 & 0.008 \\
    \bottomrule
  \end{tabular}
\end{table}

\begin{figure}[t]
  \centering
  \includegraphics[width=\linewidth]{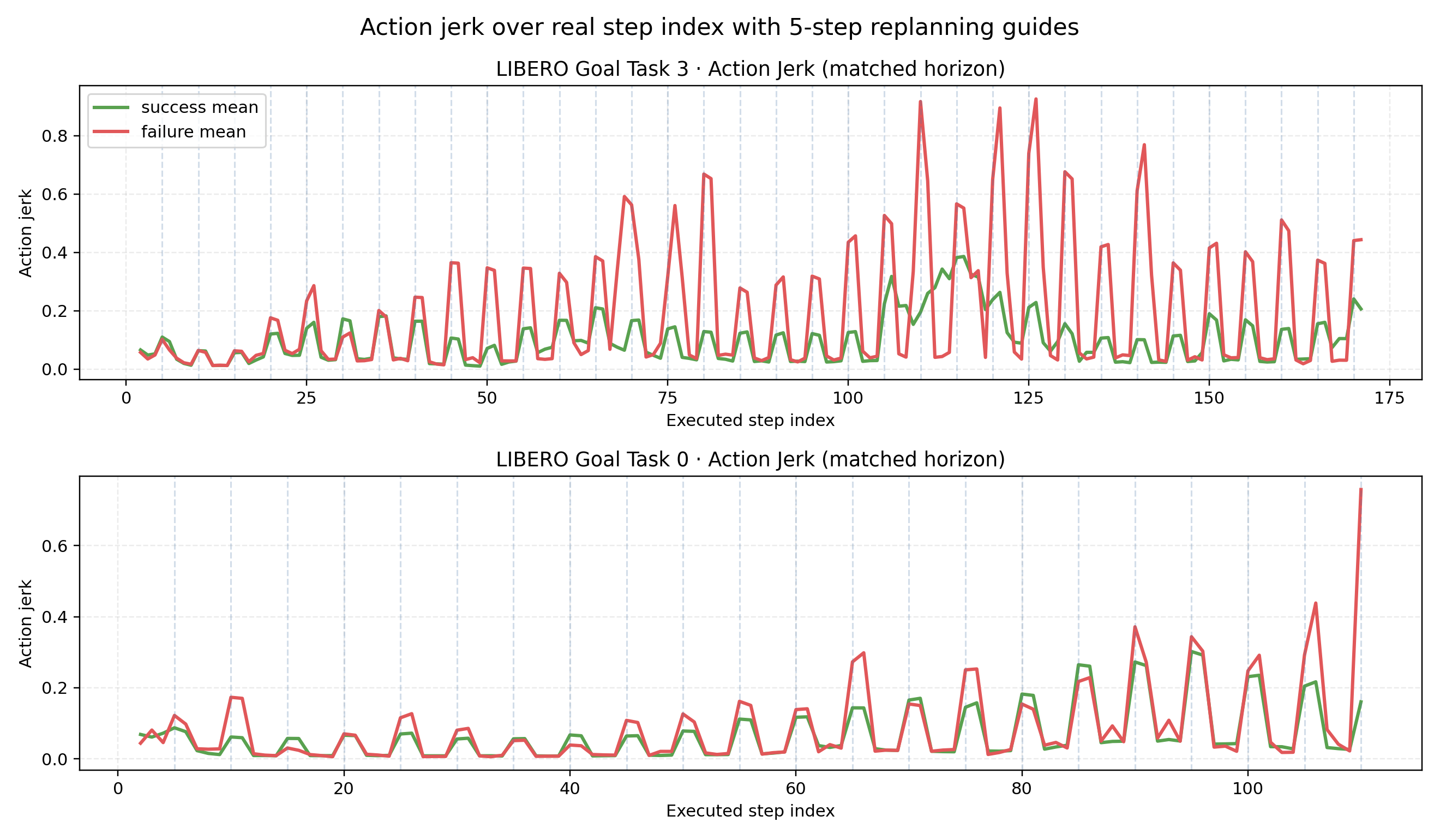}
  \caption{Matched-horizon action-jerk time courses. Even after truncating successful and failed rollouts to a common horizon, failed trajectories still exhibit stronger jerk pulses around replanning boundaries. Vertical dashed lines mark boundaries every five steps.}
  \label{fig:matched_horizon_en_v3}
\end{figure}

\subsection{Noise controllability under fixed contexts}

Outcome-level separation alone is not enough; artifact must also vary with noise under a fixed context. We therefore hold the observation context fixed and vary only latent noise. On goal 3, we sample 24 noise values for each of 16 contexts, yielding 384 samples in total. The mean context-wise standard deviation is $\sigma = 0.040$ for local boundary gap and $\sigma = 0.038$ for boundary transition jerk; the largest context-level standard deviations reach $0.097$ and $0.084$. This already rules out a simple explanation in which large artifact is merely a by-product of context change. Instead, the data suggest that context determines the region of artifact space, while noise determines which variations within that region can be pushed consistently. That \emph{steerable component} is the object of study here.

Because boundary artifact arises from the transition between two adjacent chunks, Appendix~\cref{tab:appendix_noise_decomp_en_v3} further decomposes the contributions of the current-chunk noise $\noise_0$ and the next-chunk noise $\noise_1$. Both materially affect the first-boundary artifact, but the dominant term is not fixed; it varies with task and context. A representative slice of the underlying context--noise gap matrix used in this fixed-context analysis is shown in Appendix~\cref{tab:appendix_context_noise_slice_en_v3}.

\begin{figure}[t]
  \centering
  \includegraphics[width=0.82\linewidth]{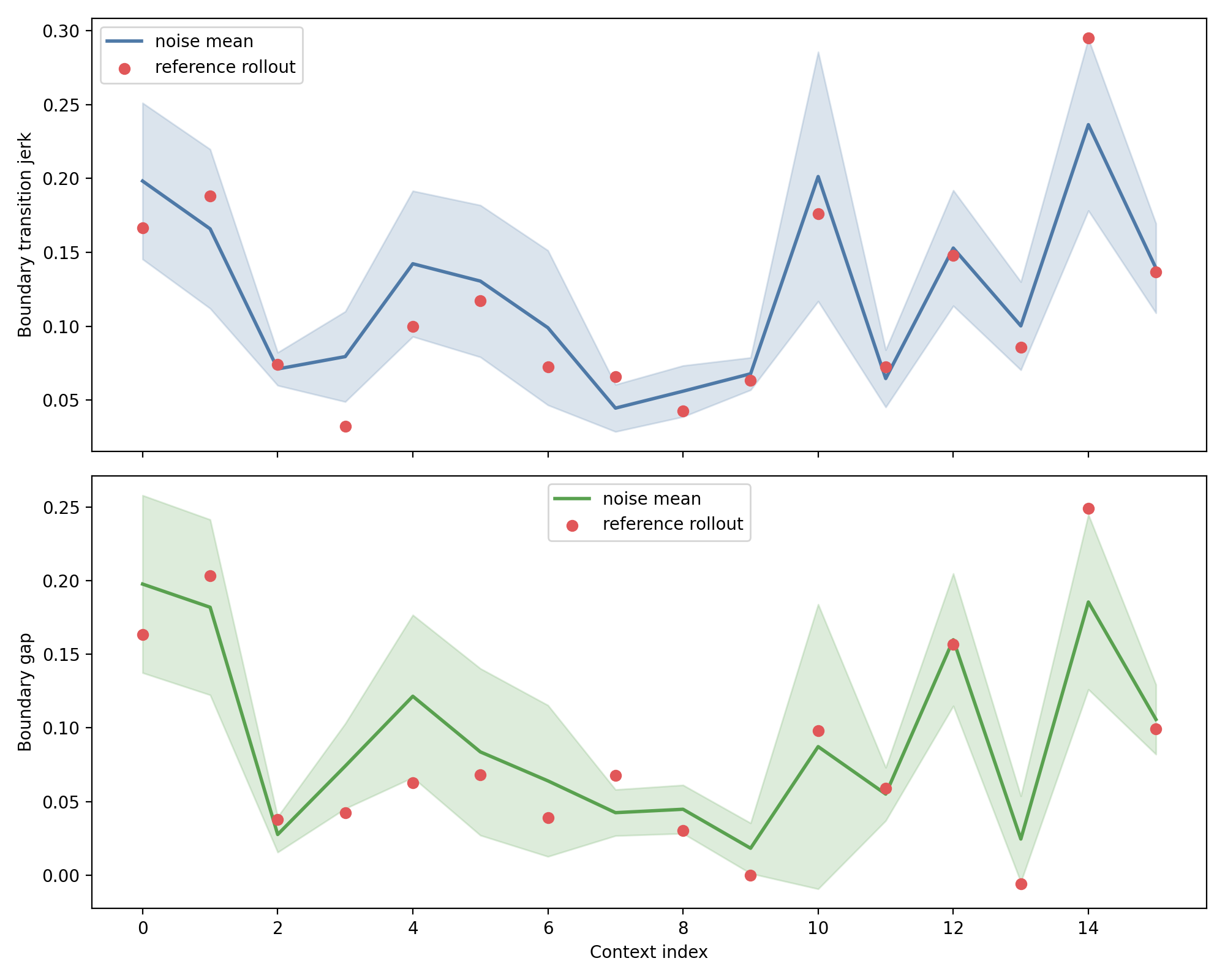}
  \caption{Artifact variation when the observation context is fixed and only latent noise is changed. Shaded regions indicate mean $\pm$ standard deviation across noise samples under the same context, and red dots denote reference rollouts. Top: boundary transition jerk. Bottom: local boundary--interior jerk contrast.}
  \label{fig:noise_scan_en_v3}
\end{figure}

\subsection{Local controllability along noise directions}

Noise-dependent variation under a fixed context still does not show that artifact can be pushed along a \emph{signed direction}. We therefore search, at each fixed observation context, for a local noise direction $\mathbf{d}$ and write the perturbed noise as
\[
  \noise'=\noise+\alpha \mathbf{d},
\]
where $\alpha$ is the signed steering strength along that direction. Concretely, we sample $12$ random unit directions per context, probe each at $\alpha=\pm\epsilon$ with $\epsilon=0.5$, and select the direction whose first-boundary transition jerk shows the largest separation between the $+\epsilon$ and $-\epsilon$ probes. We then sweep $\alpha$ along the selected direction. Across model checkpoints, direction magnitudes are normalized by $\sqrt{\dim(\noise)}$ so that the per-step perturbation scale is comparable. On \texttt{pi0.5-libero} for LIBERO-10 task 8, the resulting local signed response is strong on this small panel: across 4 fixed contexts, $\alpha$ has mean Pearson correlation $0.904$ with first-boundary transition jerk and $0.968$ with first-boundary gap; the mean gap range is $0.108$, already comparable in scale to the success/failure group difference.

\begin{figure}[t]
  \centering
  \includegraphics[width=0.96\linewidth]{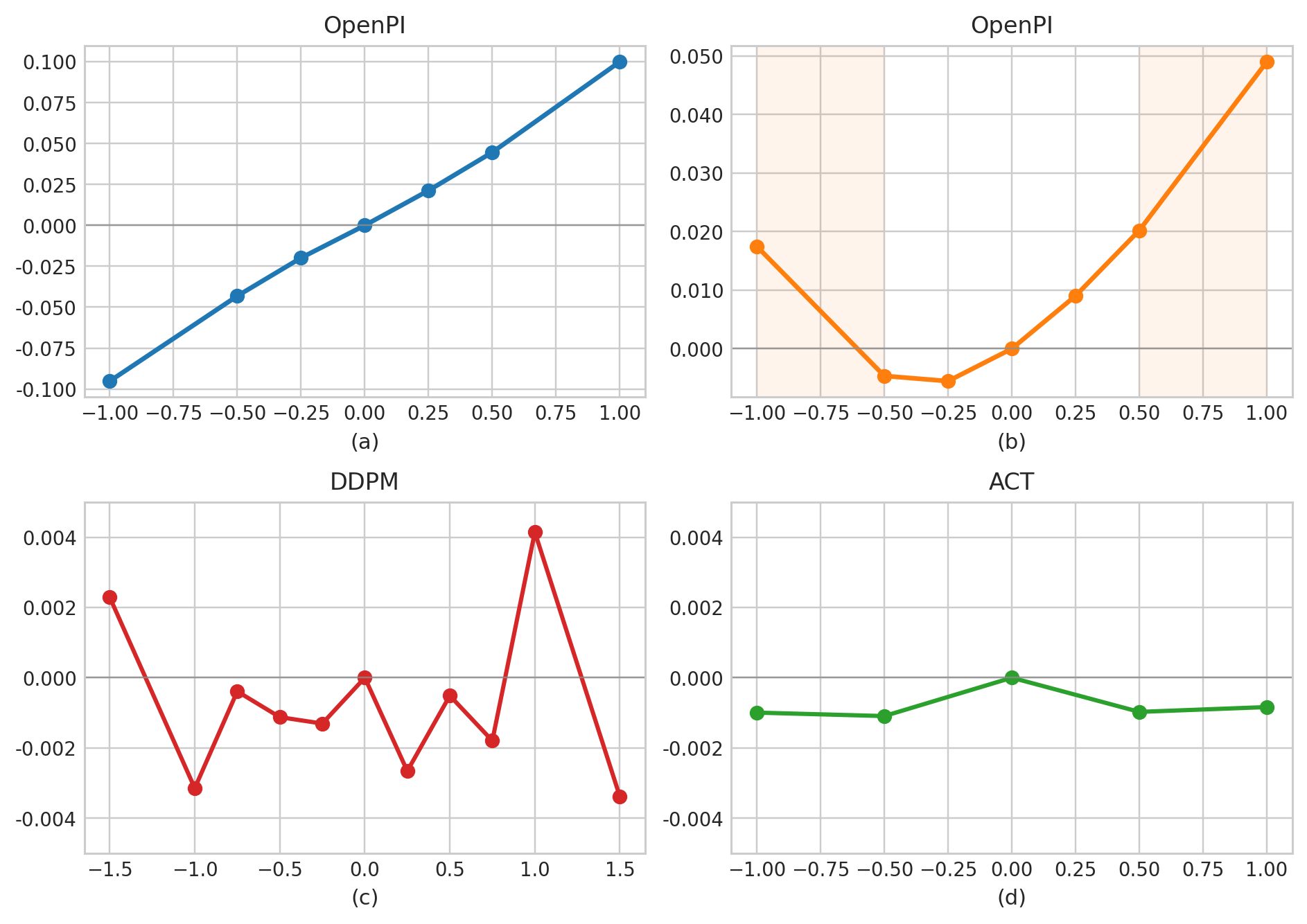}
  \caption{Local controllability along noise directions and information-path conditions. The horizontal axis is $\alpha$ in all four panels, and each vertical axis is centered at $\alpha=0$, so the comparison is about response shape rather than absolute scale. (a) OpenPI with a near-linear response; (b) OpenPI bending at large $|\alpha|$; (c) DDPM collapse; (d) ACT remaining nearly flat.}
  \label{fig:directional_bridge_en_v3}
\end{figure}

As shown in \Cref{fig:directional_bridge_en_v3}, the top row gives two representative fixed-context sweeps. Panel (a) is close to affine across the scan interval: first-boundary gap rises steadily as $\alpha$ increases. Panel (b) shows why this result should be interpreted as a \emph{local operating regime}, rather than a global linear law valid at arbitrary steering strength: once $|\alpha|$ approaches the edge of the scan interval, the response begins to bend. This also explains why the later interventions use moderate steering strengths instead of simply increasing $\alpha$ without bound; overly strong steering tends to move sampling away from the model's usual local noise regime and thereby weakens interpretability and stability.

\paragraph{Real-robot directional response.}
We also observe the same signed local response on a real Aubo i5 setup.
Starting from a single fixed real-robot execution context, sweeping the steering strength $\alpha$ along the selected OpenPI noise direction produces a monotone change in the centered first-boundary gap, with Pearson $r=0.968$ and $R^2=0.938$.
This result is used as a hardware-side check of the local noise-to-artifact response, rather than as an outcome-level success claim. A broader hardware-side replication across additional contexts is left to future work given the present access constraints to the physical platform.

\begin{figure}[t]
  \centering
  \includegraphics[width=0.82\linewidth]{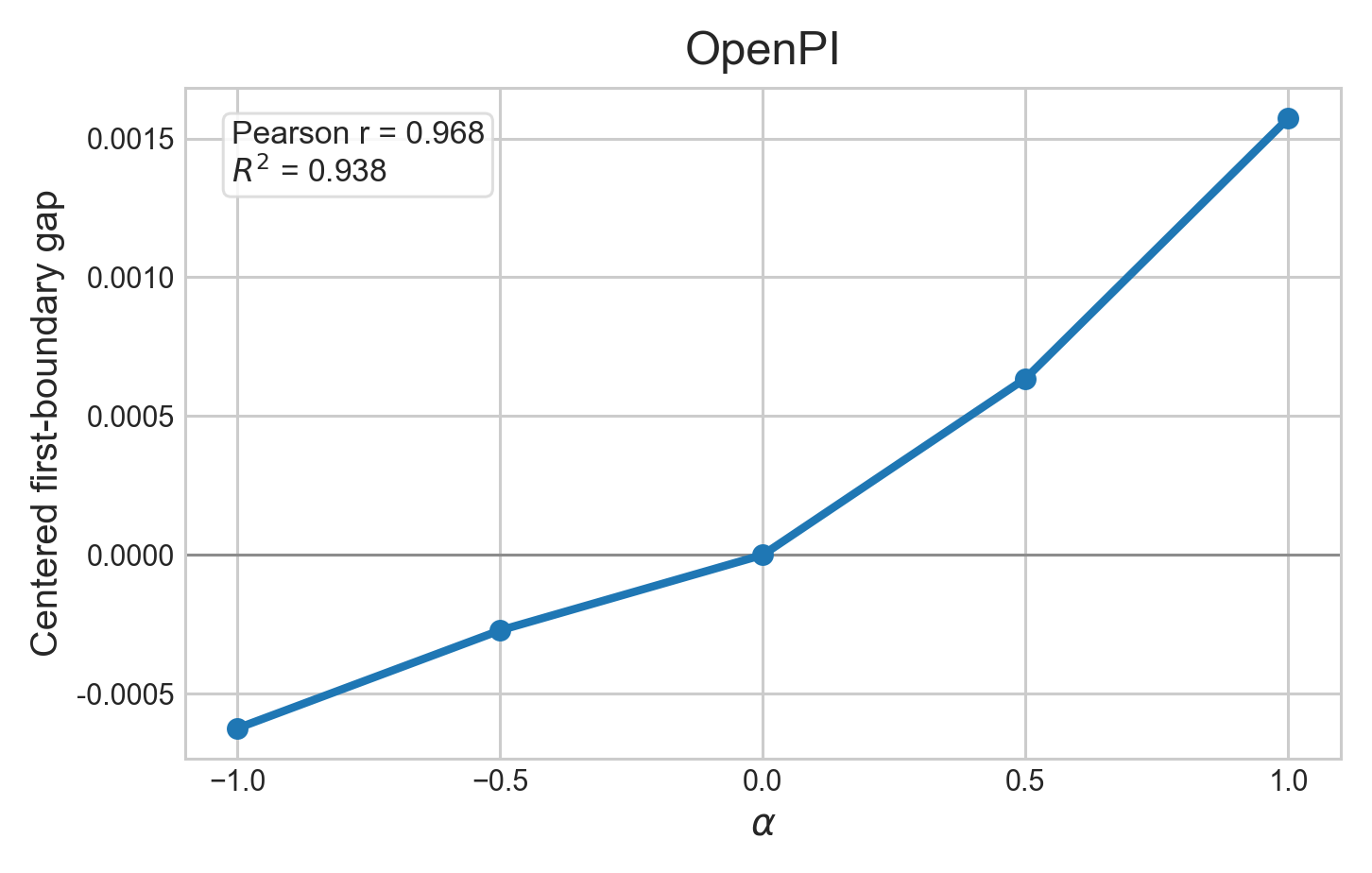}
  \caption{Real-robot directional scan on an Aubo i5 platform. Under a fixed execution context, sweeping $\alpha$ along an OpenPI noise direction produces a monotone response in the centered first-boundary gap. The result supports hardware-side local controllability of chunk-boundary artifact, without making an outcome-level success claim.}
  \label{fig:auboi5_real_robot_directional_response_en_v3}
\end{figure}

\subsection{Information-path condition: ACT and DDPM/DDIM}

The previous subsection shows that OpenPI / flow matching admits stable controllability along local noise directions. The next question is whether this is a generic property of stochastic chunked policies, or whether it depends on an intact transmission path from initial noise to action output. We therefore use two complementary controls. ACT serves as a deterministic reference: it shows only weak residual artifact and is almost unresponsive to signed latent perturbations, consistent with the near-flat response in \Cref{fig:directional_bridge_en_v3}(d). We also keep Diffusion Policy weights fixed and change only the sampling rule. Under the default DDPM sampler, the mean Pearson correlation across 8 fixed contexts is only $0.114$ for first-boundary gap and $0.052$ for first-boundary transition jerk. Even setting the variance of subsequent DDPM reverse steps to zero does not restore controllability: the mean correlations remain $0.114$ and $0.052$, and across these 8 contexts not a single one reaches even a stable linear interval of $|\alpha| \ge 0.25$ under $R^2 \ge 0.95$. By contrast, switching to DDIM ($\eta=0$) raises the same correlations to $0.961$ and $0.927$ and restores a wide, near-linear signed response regime in most contexts.

Fresh randomness at each DDPM step is not enough to explain the loss of steering. If reverse-step random reinjection were the whole story, then removing the later-step variance should restore the local directional structure; it does not. The reason is that zero-variance DDPM only replaces later stochastic sampling with its reverse-step mean; it does not turn DDPM into DDIM. The original Markovian reverse updates still remix and distort the local signed direction carried by the initial noise as denoising proceeds. Determinism alone is therefore not sufficient. What matters is whether the sampling rule preserves a \emph{stable transmission path} from initial noise to final action output. Only DDIM restores a usable directional manifold under the same weights. Chunk-boundary artifact is therefore not a purely execution-side random error. It is a structure variable in noise space that can be driven directionally. \Cref{tab:mechanism_summary_en_v3} summarizes this mechanism-level comparison across policy states.

\begin{table}[t]
  \centering
  \small
  \caption{Mechanism-level comparison across policy states. The point is not direct comparability of absolute values across tasks, but the pattern: stochastic policies exhibit large and controllable artifact, the deterministic reference exhibits only weak residual artifact, and steering effectiveness depends on whether the noise pathway is accessible.}
  \label{tab:mechanism_summary_en_v3}
  \setlength{\tabcolsep}{4pt}
  \renewcommand{\arraystretch}{1.06}
  \resizebox{\linewidth}{!}{%
  \begin{tabular}{@{}llll@{}}
    \toprule
    Policy state & Representative metric & Steering signal & Interpretation \\
    \midrule
    ACT & first-boundary gap $\approx 0.009$ & no stable signed separation & weak residual artifact \\
    OpenPI (flow matching; main experiments use \texttt{pi0.5-libero}) & local $\alpha$ sweep & $r_{\mathrm{gap}}/r_{\mathrm{transition}}=0.968/0.904$ & strong local directional structure \\
    Diffusion Policy + DDPM & same weights & $r_{\mathrm{gap}}/r_{\mathrm{transition}}=0.114/0.052$ & pathway largely broken \\
    Diffusion Policy + DDPM (subsequent variance set to zero) & same weights & still $0.114/0.052$ & removing reinjection does not restore directional structure \\
    Diffusion Policy + DDIM ($\eta=0$) & same weights & $r_{\mathrm{gap}}/r_{\mathrm{transition}}=0.961/0.927$ & pathway intact \\
    \bottomrule
  \end{tabular}}
\end{table}

\section{Causal Attribution of Artifact}
\label{sec:causal_attribution_en_v3}

\subsection{Full-trajectory steering and outcome change}

Once artifact becomes controllable, the next question is whether full-trajectory steering along artifact-aligned directions changes the final outcome. We use full-trajectory steering as a \emph{mechanistic intervention tool}: at each replanning boundary it suppresses or amplifies local artifact in a controlled way, so that we can observe how this mechanism variable propagates to outcome.

\begin{figure}[t]
  \centering
  \begin{subfigure}{0.48\linewidth}
    \centering
    \includegraphics[width=\linewidth]{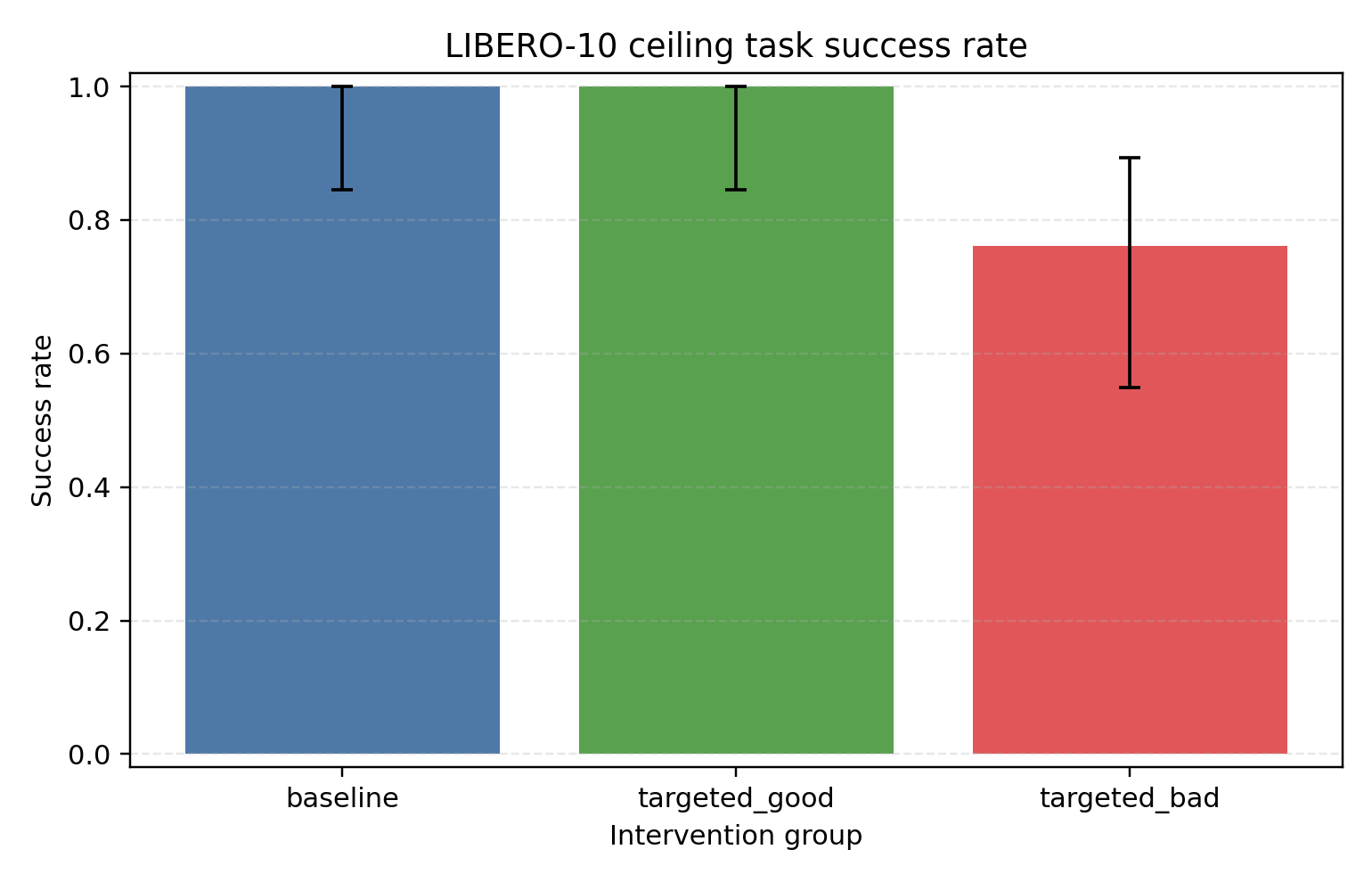}
  \end{subfigure}\hfill
  \begin{subfigure}{0.48\linewidth}
    \centering
    \includegraphics[width=\linewidth]{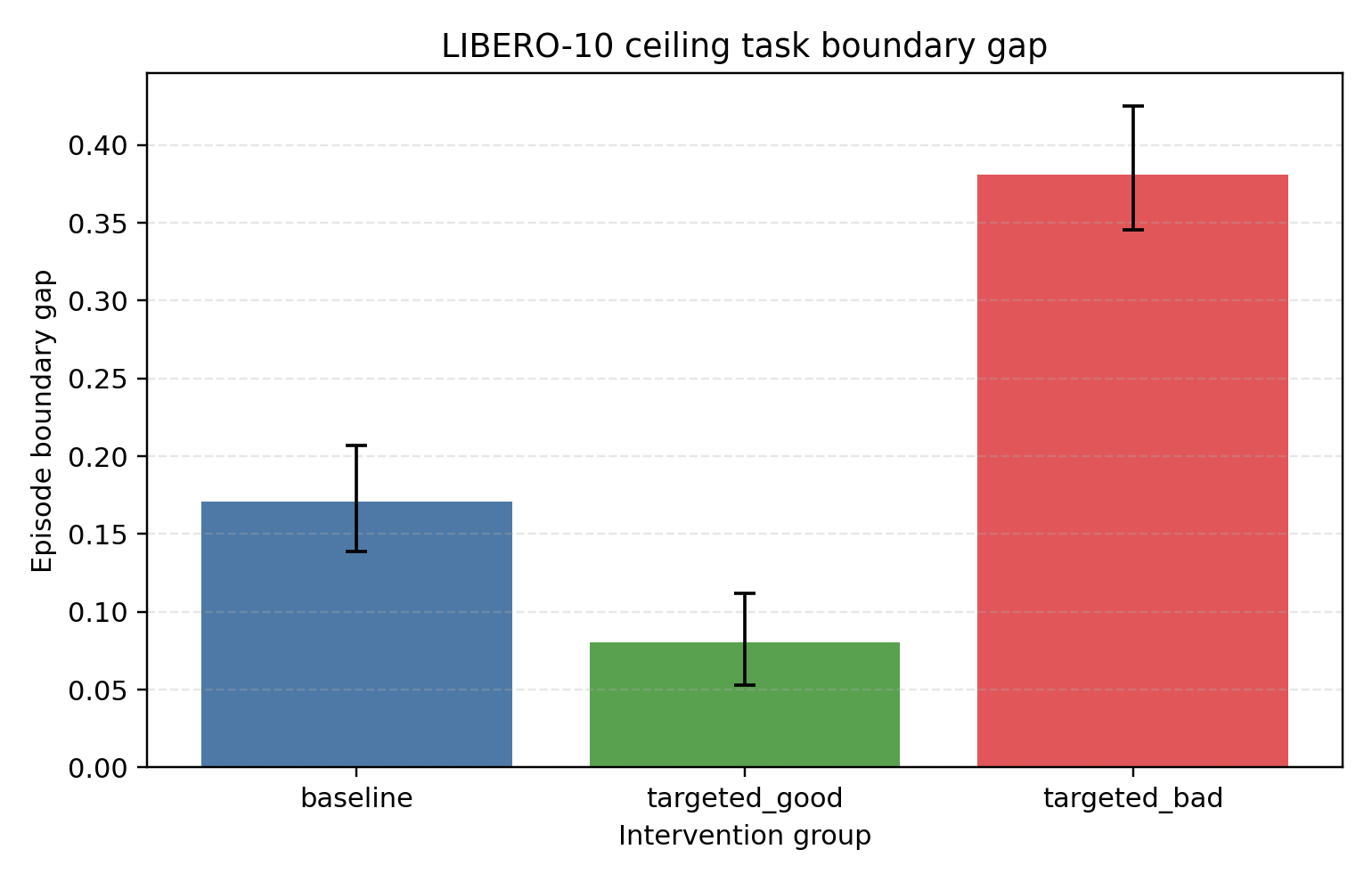}
  \end{subfigure}

  \vspace{0.75em}

  \begin{subfigure}{0.48\linewidth}
    \centering
    \includegraphics[width=\linewidth]{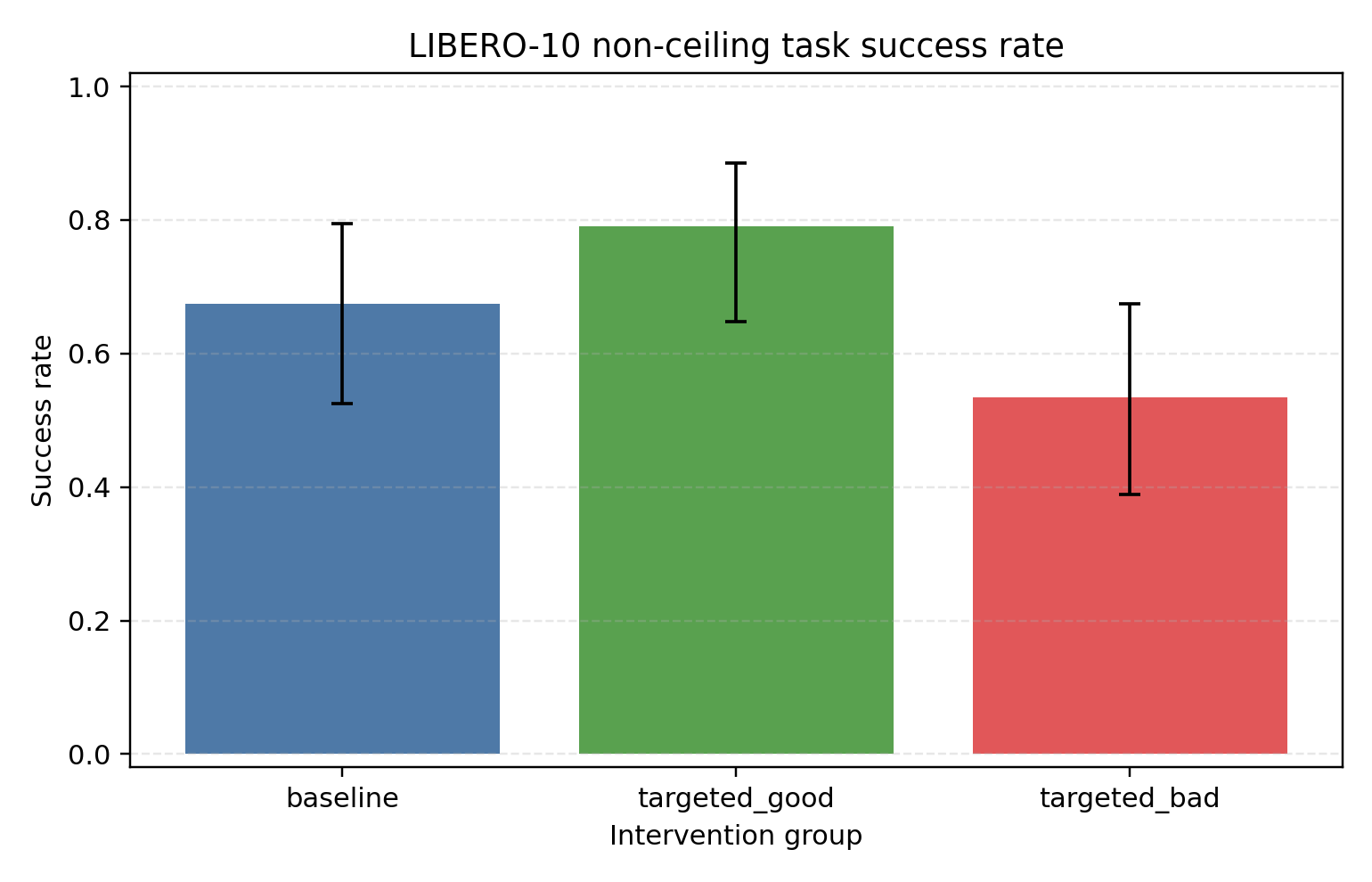}
  \end{subfigure}\hfill
  \begin{subfigure}{0.48\linewidth}
    \centering
    \includegraphics[width=\linewidth]{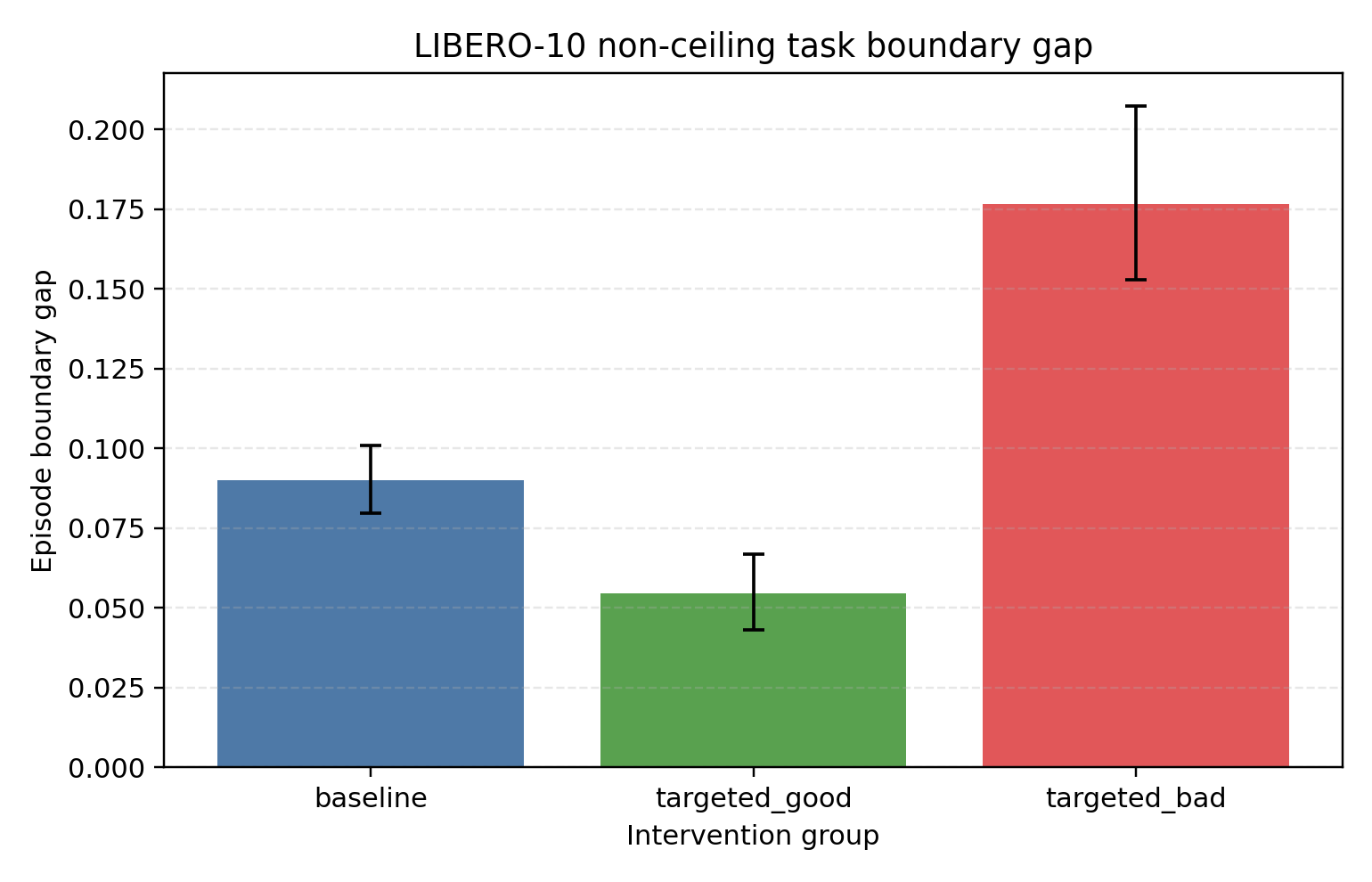}
  \end{subfigure}
  \caption{Full-trajectory steering. The top row shows a task whose baseline success rate is already near saturation: the mechanism metric is controllable, but positive success gains are compressed. The bottom row shows a task with clear room for improvement: target-low raises success from $0.674$ to $0.791$, while target-high lowers it to $0.535$; meanwhile the boundary-gap metric preserves the consistent ordering low $<$ baseline $<$ high.}
  \label{fig:fulltraj_summary_en_v3}
\end{figure}

\Cref{fig:fulltraj_summary_en_v3} shows two complementary cases. Target-low and target-high denote steering along previously calibrated low-artifact and high-artifact directions. We keep only the two results that illustrate the point most directly: one near-saturation task, used to show that a success ceiling compresses positive gains, and one non-saturated task 8 result, used to show that the outcome distribution moves together with artifact when there is still room for improvement. The remaining pooled results are shown in Appendix~\Cref{fig:appendix_fulltraj_overview_en_v3}.

In the first type of task, baseline success is already close to saturation. This pooled comparison uses $n=21$ episodes per group: both baseline and target-low achieve success rate $1.0$, whereas target-high drops to $0.762$, i.e., $23.8$ percentage points below baseline. The corresponding mean boundary-gap values move from $0.171$ down to $0.081$ or up to $0.381$. With $n=21$ and baseline already at $1.0$, this comparison cannot resolve a positive uplift on this task; it is reported only to illustrate the ceiling-compression effect on the success-rate side while the boundary-gap metric remains responsive in both directions.

The second type of task still has substantial room for improvement. On task 8, each group contains $n=43$ episodes, and the success ordering
\[
  0.791 \;>\; 0.674 \;>\; 0.535
\]
exactly matches the boundary-gap ordering
\[
  0.0546 \;<\; 0.0899 \;<\; 0.1765.
\]
Target-low improves success by $11.7$ percentage points relative to baseline, whereas target-high decreases it by $13.9$ points. The $95\%$ confidence intervals of the three success rates are $[0.648,\,0.886]$, $[0.525,\,0.795]$, and $[0.389,\,0.675]$, respectively. This indicates that full-trajectory steering is acting on a mechanism variable that materially affects failure risk.

\subsection{Direction reversal under fixed key contexts}

The strongest bridge evidence comes from paired interventions at fixed key contexts. Concretely, we fix a rollout prefix up to a specified replanning boundary, treat the resulting local execution state as one context instance, and then change only the boundary noise or steering direction while keeping the prefix and pairwise matched future continuation unchanged. In several tightly controlled contexts, the preferred artifact direction lines up cleanly with outcome change: some key contexts favor lower artifact, whereas others favor higher artifact. ``Artifact affects outcome'' is therefore not a globally monotone law across contexts, but a strongly context-dependent local regularity.

These key contexts are not two identifiers picked post hoc from an unconstrained search space. The selection pipeline has three stages. First, we run a cross-index scan over rollouts on task 8 to enumerate candidate key contexts; for the high-artifact-favoring side we additionally restrict to contexts with $\text{success}_{\text{high}}=1$, $\text{success}_{\text{baseline}}=0$, $\text{success}_{\text{low}}=0$, and $\text{gap}_{\text{high}}>\text{gap}_{\text{baseline}}>\text{gap}_{\text{low}}$, which yields $223$ candidates. Second, each candidate passes through a local controllability check that retains only contexts with monotone $\alpha$-response. Third, surviving candidates are evaluated with matched-continuation paired validation, with pilot $\alpha$-selection and held-out evaluation on disjoint matched-continuation banks so that the reported uplift is not the result of $\alpha$ tuning on the same data. The two rows in \Cref{tab:fixed_key_contexts_en_v3} are the strongest survivors of this pipeline in the low-favoring and high-favoring directions, respectively. Several other high-artifact candidates that passed local screening (e.g., $(10,7,20)$, $(10,16,35)$, $(10,29,37)$) yielded null or marginal effects under the same matched-continuation protocol, which is why direction preference cannot be claimed at the task level. We also explicitly checked that the ctx $(10,24,21)$ uplift is not driven by future-continuation seed luck: under pairwise-matched future continuations and an identical prefix, the only quantity that changes is the boundary noise itself, so the uplift is attributable to the local first-boundary artifact change it induces.

The two paired results in \Cref{tab:fixed_key_contexts_en_v3} both come from task 8, yet the optimal direction is opposite. The first key context favors target-low, whereas the second favors target-high. In counts, the first changes from $127/180$ to $147/180$, while the second changes from $2/60$ to $43/60$. More importantly, future continuation is pairwise matched in these designs, so the outcome shift cannot be attributed to downstream rollout randomness; it must arise from the current boundary noise and the local artifact state that it induces.

\begin{table}[t]
  \centering
  \scriptsize
  \setlength{\tabcolsep}{4pt}
  \caption{Paired steering results for two fixed key contexts within task 8. Both come from the same top-12 candidate list under matched-continuation validation, but their optimal intervention directions are opposite. The first row aggregates $n=180$ matched pairs and the second aggregates $n=60$ matched pairs. The baseline in the second row is the conditional success rate under the frozen reference noise at that key context, not the task-level average success rate.}
  \label{tab:fixed_key_contexts_en_v3}
  \begin{tabular}{lcccccc}
    \toprule
    Context $(seed,trial,replan)$ & Preferred direction & Pairs & Baseline & Steered & Uplift & $p$ \\
    \midrule
    $(10,23,19)$ & target-low & 180 & $0.706$ & $0.817$ & $+0.111$ & $0.0093$ \\
    $(10,24,21)$ & target-high & 60 & $0.033$ & $0.717$ & $+0.683$ & $4.55\times 10^{-13}$ \\
    \bottomrule
  \end{tabular}
\end{table}

\subsection{Context dependence of the optimal steering target}

The larger fixed-context extension directly rejects the idea that the optimal target is a global constant. Here we share the same set of 16 candidate noises across 192 fixed contexts, yielding 3072 context-noise combinations, and ask a simple question: does the same noise remain best across contexts?

\Cref{tab:adaptive_interaction_en_v3} shows that, after subtracting the baseline gap each context carries under its reference noise, the remaining variation is split almost evenly between the context main effect and the context$\times$noise interaction term. At the same time, the globally best low-artifact noise remains locally optimal for only $3.1\%$ of contexts. In other words, that globally best noise is \emph{not} the best choice for $96.9\%$ of contexts. By contrast, if we allow context-wise selection within the same 16-noise candidate set, the mean gap drops from $0.087$ to $0.005$.

The quantity that must be predicted is therefore not a globally reusable high/low artifact constant, but a control target modulated by context. This also explains why adaptive steering is difficult to stabilize: the problem is not regression to a single low-dimensional target, but a strongly context-dependent local decision problem.

\begin{table}[t]
  \centering
  \scriptsize
  \setlength{\tabcolsep}{5pt}
  \caption{Quantitative results from the larger fixed-context extension on task 8. The analysis uses 192 contexts and a shared set of 16 candidate noises to quantify the context$\times$noise interaction and to show why the optimal steering target cannot be treated as a global constant across contexts.}
  \label{tab:adaptive_interaction_en_v3}
  \begin{tabular}{@{}p{0.36\linewidth}p{0.56\linewidth}@{}}
    \toprule
    Statistic & Value \\
    \midrule
    Variance decomposition of raw gap & context main effect $59.1\%$; noise main effect $1.4\%$; context$\times$noise interaction $39.4\%$ \\
    Residual decomposition after subtracting baseline gap & context main effect $49.1\%$; noise main effect $1.8\%$; context$\times$noise interaction $49.1\%$ \\
    Fraction of contexts where the globally best low-artifact noise remains locally optimal & $3.1\%$ \\
    Mean gap under the globally best low-artifact noise & $0.087$ \\
    Mean gap under context-wise optimal noise selection & $0.005$ \\
    Relative adaptive gain & $93.8\%$ reduction in mean gap \\
    \bottomrule
  \end{tabular}
\end{table}

\section{Discussion}
\label{sec:discussion_en_v3}

\paragraph{Why the mechanism evidence is stronger than the method-level closure.}
The most stable evidence in this paper remains at the mechanism level, not at the level of a unified online controller. Once these local regularities are compressed into a single controller, candidate-set size, label noise, local credit assignment, and sequential intervention budget all enter the problem at once. Current diagnostics indicate that the first-stage question---which contexts are worth intervening on---is not entirely unlearnable, but the real bottleneck lies in the second-stage dispatch: the controller must decide when to intervene, whether to choose low or high, and how to allocate a limited intervention budget. More fundamentally, direction preference itself can reverse within the same task. On task 8, some complete local contexts favor low artifact, some favor high artifact, and many are best left unperturbed, so there is no single fixed direction that transfers across contexts. The current adaptive-steering results therefore show only local positive signal, not yet stable online gains.

\paragraph{Integrity of the information path.}
The DDPM/DDIM comparison shows that what determines steering success is not stochasticity per se, but whether the control signal can be transmitted stably from the initial noise to the action output. Zero-variance DDPM still fails to recover directional structure, further excluding the simple explanation that the problem is merely per-step random-noise reinjection. As generative robot policies increasingly move toward few-step paradigms, such as flow matching and consistency-style models, this information-path issue is likely to become even more exposed.

\paragraph{Online cost and future method directions.}
The current intervention procedure still depends on local direction search, so its online cost remains well above that of a lightweight real-time module. If this line is pursued further, a budget-aware context-adaptive controller is a more sensible target than continued tuning of the current gate. Before that, the more realistic intermediate path remains staged key-context screening followed by local branch validation.

\section{Conclusion}
\label{sec:conclusion_en_v3}

This paper provides a mechanism-level attribution of chunk-boundary artifact in action-chunked policies by tracing the chain from noise to artifact to outcome. The experiments show that stochastic action-chunked policies contain structures in noise space that can stably drive boundary artifact; that the DDPM, zero-variance DDPM, and DDIM comparison ties this controllability to the integrity of the information path from initial noise to action output; that directional intervention on this structure at fixed key contexts can significantly change task outcome; and that fixed-context oracle analysis shows the optimal steering target is determined by context rather than by a global constant shared across contexts. Chunk-boundary artifact should therefore not be treated as a mere execution disturbance. It is a variable that can be analyzed, controlled, and mechanistically linked to task outcome. This clarifies the source of boundary problems in action-chunked policies and provides a concrete starting point for future context-adaptive steering.

\clearpage
\appendix
\setcounter{figure}{0}
\renewcommand{\thefigure}{A\arabic{figure}}
\setcounter{table}{0}
\renewcommand{\thetable}{A\arabic{table}}

\begin{center}
{\LARGE\bfseries Appendix}
\end{center}
\vspace{0.5em}

\subsection*{Appendix A: Context--Noise Matrix Slice}

\begin{table}[H]
  \centering
  \scriptsize
  \setlength{\tabcolsep}{5pt}
  \caption{Top-left $6 \times 8$ slice of the OpenPI context--noise boundary-gap matrix used in the fixed-context noise analysis. Columns correspond to contexts $c1$ through $c6$, and rows correspond to noise samples $z1$ through $z8$.}
  \label{tab:appendix_context_noise_slice_en_v3}
  \begin{tabular}{lrrrrrr}
    \toprule
    Noise / Context & c1 & c2 & c3 & c4 & c5 & c6 \\
    \midrule
    z1 & 0.199 & 0.166 & 0.029 & 0.044 & 0.105 & 0.102 \\
    z2 & 0.169 & 0.158 & 0.018 & 0.067 & 0.126 & 0.061 \\
    z3 & 0.169 & 0.183 & 0.012 & 0.073 & 0.112 & 0.065 \\
    z4 & 0.208 & 0.212 & 0.011 & 0.073 & 0.103 & 0.047 \\
    z5 & 0.109 & 0.096 & 0.046 & 0.080 & 0.119 & 0.142 \\
    z6 & 0.224 & 0.249 & 0.024 & 0.059 & 0.055 & 0.050 \\
    z7 & 0.169 & 0.150 & 0.032 & 0.070 & 0.148 & 0.110 \\
    z8 & 0.255 & 0.280 & 0.044 & 0.087 & 0.009 & 0.020 \\
    \bottomrule
  \end{tabular}
\end{table}

\subsection*{Appendix B: $\noise_0$ / $\noise_1$ Decomposition}

\begin{table}[H]
  \centering
  \small
  \setlength{\tabcolsep}{6pt}
  \caption{First-boundary noise decomposition. We vary only current-chunk noise $\noise_0$, only next-chunk noise $\noise_1$, or both together, and report the standard deviation of the first-boundary artifact metrics. BTJ denotes first-boundary transition jerk. Each condition uses 2 contexts $\times$ 5 samples; goal 0 is overall less noise-sensitive than goal 3 ($\sim 0.02$ vs $\sim 0.03$), and at this sample size the goal 0 ``vary both'' std is not separable from the single-factor conditions.}
  \label{tab:appendix_noise_decomp_en_v3}
  \begin{tabular}{llcc}
    \toprule
    Task & Condition & BTJ std & gap std \\
    \midrule
    \multirow{3}{*}{goal 3}
      & vary $\noise_0$ only & 0.029 & 0.030 \\
      & vary $\noise_1$ only & 0.035 & 0.040 \\
      & vary both & 0.038 & 0.038 \\
    \midrule
    \multirow{3}{*}{goal 0}
      & vary $\noise_0$ only & 0.022 & 0.020 \\
      & vary $\noise_1$ only & 0.024 & 0.025 \\
      & vary both & 0.017 & 0.019 \\
    \bottomrule
  \end{tabular}
\end{table}

This decomposition shows that chunk noise on both sides of the boundary contributes to first-boundary artifact. For goal 3, $\noise_1$ contributes slightly more than $\noise_0$, which matches the intuition that the artifact depends strongly on what the next chunk generates. But this relative ordering should not be read as a global law: in a harder task-8 replication, $\noise_0$ is slightly stronger than $\noise_1$. The more robust conclusion is therefore that \emph{both sides of the boundary matter, and the dominant term depends on task and context.}

\subsection*{Appendix C: Full-Trajectory Steering Summary}

\begin{figure}[H]
  \centering
  \begin{subfigure}{0.48\linewidth}
    \centering
    \includegraphics[width=\linewidth]{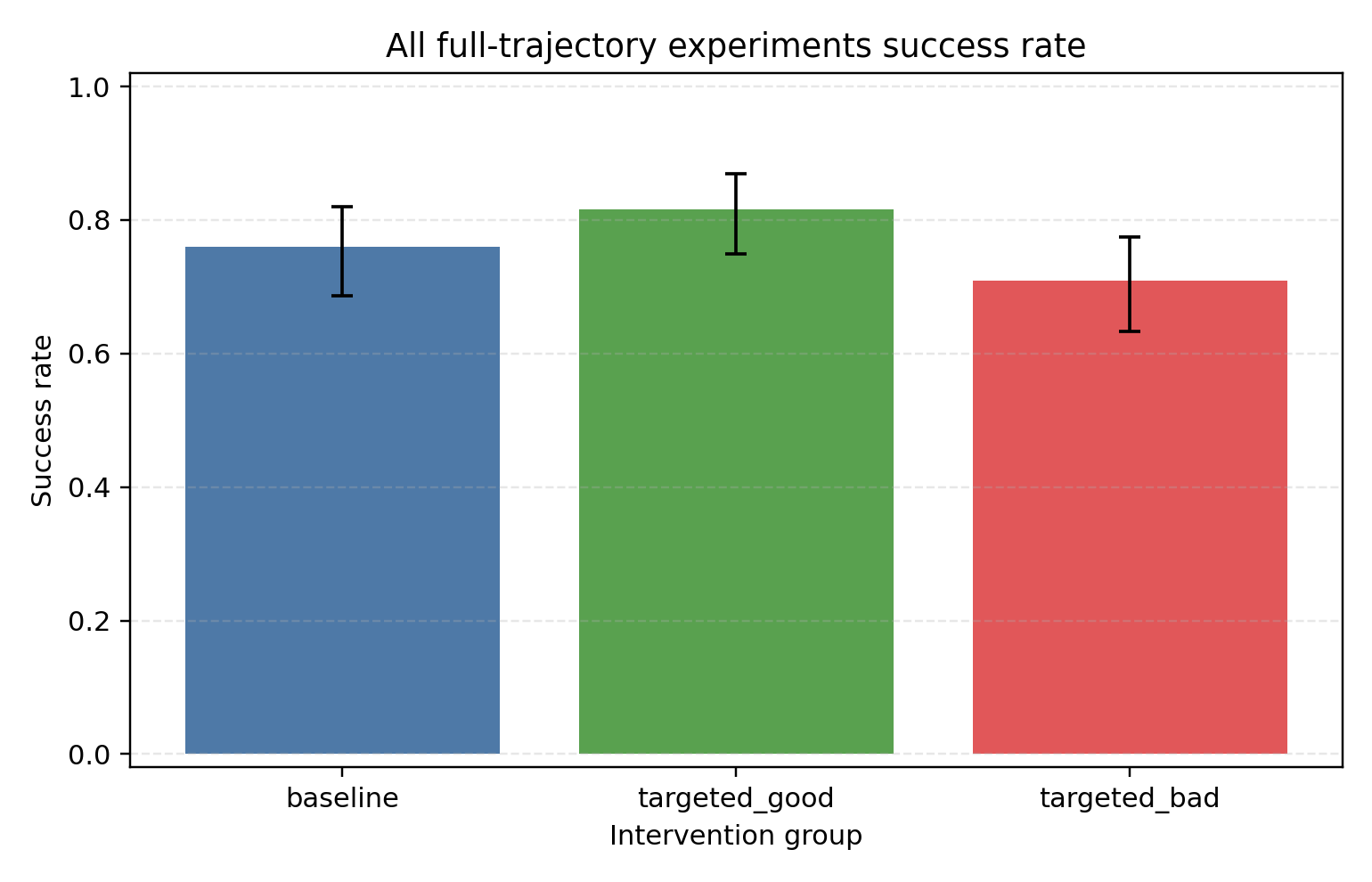}
  \end{subfigure}\hfill
  \begin{subfigure}{0.48\linewidth}
    \centering
    \includegraphics[width=\linewidth]{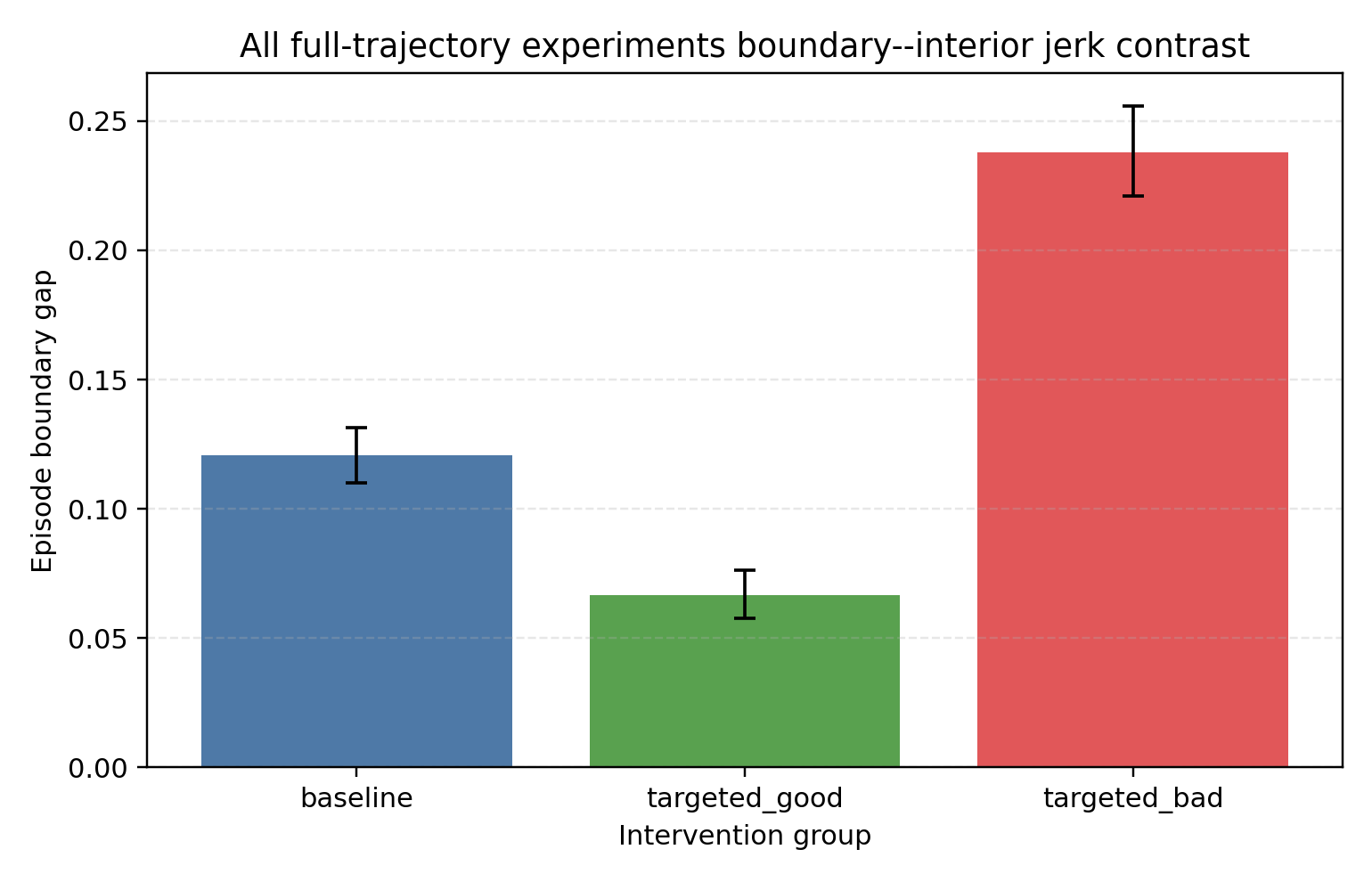}
  \end{subfigure}
  \caption{Aggregate summary of the full-trajectory steering experiments ($n=158$ per group). Left: success rate. Right: episode-level boundary--interior jerk contrast. The aggregate mixes near-saturation baselines, settings with substantial room for improvement, settings close to a failure floor, and settings with still-limited sample size; accordingly, pooled success-rate separation is compressed, while the directional ordering of the mechanism metric is preserved.}
  \label{fig:appendix_fulltraj_overview_en_v3}
\end{figure}

\subsection*{Appendix D: Adaptive Steering Exploration}

\begin{table}[H]
  \centering
  \small
  \setlength{\tabcolsep}{5pt}
  \caption{Supplementary diagnostics for adaptive steering. These are not presented as positive method results, but as key diagnostics: whether first-stage sensitivity screening is learnable, whether the online controller exhibits local positive signal, and whether adaptive selection becomes feasible at the artifact level before it becomes stable at the outcome level.}
  \label{tab:appendix_adaptive_diag_en_v3}
  \resizebox{\linewidth}{!}{%
  \begin{tabular}{lp{0.62\linewidth}}
    \toprule
    Component & Result \\
    \midrule
    First-stage sensitivity predictor & A linear classifier using only the current context predicts whether a local context is intervention-worthy, i.e.\ whether offline fixed-context validation shows that some intervention branch outperforms baseline there; under leave-one-trial-out evaluation it reaches accuracy $0.651 \pm 0.017$ and AUROC $0.593 \pm 0.040$, above the majority-class baseline accuracy of $0.617$ \\
    After connecting the first-stage filter to the online controller & the number of interventions drops to roughly $1/3$--$1/4$ of the original level, but after aggregation over a longer window success drops from the baseline $0.6$ to $0.2$, indicating that once the filter fires, second-stage dispatch---when to intervene, which direction to choose, and how to allocate budget---remains the main bottleneck \\
    Online controller & a local positive result appears on a shorter evaluation window with $0.4 \rightarrow 0.8$, but after aggregation over a longer window it becomes only $0.5 \rightarrow 0.6$; the one-sided sign test against baseline gives $p=0.5$ \\
    Fixed-context predictor with 16 shared candidate noises & single-model selection reduces held-out mean gap from $0.1068$ to $0.0815$; the validation-selected top-3 ensemble yields positive gain on $5/5$ splits, with mean absolute gap gain $0.0246$ \\
    \bottomrule
  \end{tabular}}
\end{table}

The outcome-level controller is not yet stable, but artifact-level adaptation is already learnable. Under a fixed-context setting with 16 shared candidate noises, we train a context-adaptive predictor that directly predicts which candidate noise is more suitable for the current context. On held-out contexts, the single-model result reduces predictor-selected mean gap from the global fixed low-noise baseline of $0.1068$ to $0.0815$, a relative reduction of $23.7\%$. Across 5 logical-context splits, the validation-selected top-3 ensemble achieves positive gains on all 5 splits, with mean absolute gap gain $0.0246$.

\emph{Adaptive selection is thus already feasible at the artifact level; the remaining difficulty is to carry that advantage through to the outcome level in a stable way.}

\begin{table}[H]
  \centering
  \small
  \setlength{\tabcolsep}{5pt}
  \caption{Results for the fixed-context predictor with 16 shared candidate noises. Single model denotes the best single model chosen by validation within each split; top-3 ensemble denotes the ensemble of the three models ranked highest by validation performance.}
  \label{tab:appendix_bank_predictor_en_v3}
  \resizebox{\linewidth}{!}{%
  \begin{tabular}{lcccc}
    \toprule
    Setting & Held-out selected gap & Global fixed baseline gap & Absolute gain & Note \\
    \midrule
    single model (split seed 1) & 0.0815 & 0.1068 & 0.0253 & RMSE $=0.1037$, Pearson $r=0.618$ \\
    single model (5-split aggregate) & 0.0566 & 0.0763 & 0.0197 & positive gain on 4/5 splits \\
    top-3 ensemble (5-split aggregate) & 0.0517 & 0.0763 & 0.0246 & positive gain on 5/5 splits \\
    \bottomrule
  \end{tabular}}
\end{table}


\begin{thebibliography}{11}

\bibitem{black2024pi0}
K.~Black, N.~Brown, D.~Driess, A.~Escontrela, M.~Nasiriany, et~al.
\newblock $\pi_0$: A vision-language-action flow model for general robot control.
\newblock In \emph{RSS}, 2025.

\bibitem{chi2023diffusion}
C.~Chi, Z.~Xu, S.~Feng, Y.~Du, E.~Cousineau, et~al.
\newblock Diffusion Policy: Visuomotor Policy Learning via Action Diffusion.
\newblock In \emph{RSS}, 2023.

\bibitem{zhao2023aloha}
T.~Z. Zhao, V.~Kumar, S.~Levine, and C.~Finn.
\newblock Learning Fine-Grained Bimanual Manipulation with Low-Cost Hardware.
\newblock In \emph{RSS}, 2023.

\bibitem{dalva2024noiseclr}
Y.~Dalva and P.~Yanardag.
\newblock NoiseCLR: A Contrastive Learning Approach for Unsupervised Discovery of Interpretable Directions in Diffusion Models.
\newblock In \emph{CVPR}, 2024.

\bibitem{libero2023}
B.~Liu, Y.~Zhu, C.~Gao, Y.~Feng, Q.~Liu, et~al.
\newblock LIBERO: Benchmarking Knowledge Transfer for Lifelong Robot Learning.
\newblock In \emph{NeurIPS Datasets and Benchmarks}, 2023.

\bibitem{learning_native_continuation}
Y.~Liu, H.~Yu, J.~Zhao, B.~Li, D.~Zhang, M.~Li, et~al.
\newblock Learning Native Continuation for Action Chunking Flow Policies.
\newblock \emph{arXiv preprint arXiv:2602.12978}, 2026.

\bibitem{liu2025bid}
Y.~Liu, J.~I. Hamid, A.~Xie, Y.~Lee, M.~Du, and C.~Finn.
\newblock Bidirectional Decoding: Improving Action Chunking via Closed-Loop Resampling.
\newblock In \emph{ICLR}, 2025.

\bibitem{rtc2024}
K.~Black, M.~Y. Galliker, and S.~Levine.
\newblock Real-Time Execution of Action Chunking Flow Policies.
\newblock In \emph{NeurIPS}, 2025.

\bibitem{remac2025}
H.~Wang, G.~Zhang, Y.~Yan, Y.~Shang, R.~R. Kompella, and G.~Liu.
\newblock Real-Time Robot Execution with Masked Action Chunking.
\newblock In \emph{ICLR}, 2026.

\bibitem{initno2024}
X.~Guo, J.~Liu, M.~Cui, J.~Li, H.~Yang, and D.~Huang.
\newblock InitNO: Boosting Text-to-Image Diffusion Models via Initial Noise Optimization.
\newblock In \emph{CVPR}, 2024.

\bibitem{dsrl2024}
A.~Wagenmaker, M.~Nakamoto, Y.~Zhang, S.~Park, W.~Yagoub, et~al.
\newblock Steering Your Diffusion Policy with Latent-Space Reinforcement Learning.
\newblock In \emph{CoRL}, 2025.

\end{thebibliography}
\end{document}